\pdfoutput=1

\documentclass[11pt]{article}

\usepackage{acl}

\usepackage{times}
\usepackage{latexsym}
\usepackage{xspace}
\usepackage{lipsum}
\usepackage{changepage} 
\usepackage{multirow}
\usepackage{booktabs}
\usepackage{color,colortbl}
\usepackage{subcaption}
\definecolor{Gray}{gray}{0.9}
\usepackage{tabularx}
\usepackage{arydshln}
\usepackage{amsfonts}
\usepackage{amsmath}
\usepackage{amssymb}

\usepackage[T1]{fontenc}

\usepackage[utf8]{inputenc}

\usepackage{microtype}
\usepackage[textsize=scriptsize]{todonotes}

\usepackage{inconsolata}
\newcommand{\tod}{{\textsc{ToD}}\xspace}

\newcommand{\tabitem}{~~\llap{\textbullet}~~}

\usepackage{color, colortbl}

\newcommand{\method}{\textsc{SQATIN}\xspace}

\newcommand{\rparagraph}[1]{\vspace{1.6mm}\noindent\textbf{#1.}}
\newcommand{\rrparagraph}[1]{\vspace{0.4mm}\noindent\textit{#1.}}
\newcommand{\sparagraph}[1]{\vspace{0.0mm}\noindent\textbf{#1.}}

\definecolor{Gray}{gray}{0.92}
\definecolor{racing-green}{rgb}{0.0, 0.8, 0.6}
\definecolor{awesome-red}{rgb}{1.0, 0.13, 0.32}
\newcolumntype{Y}{>{\centering\arraybackslash}X}

\usepackage{todonotes}
\makeatletter
\newcommand*\iftodonotes{\if@todonotes@disabled\expandafter\@secondoftwo\else\expandafter\@firstoftwo\fi}
\makeatother

\title{\method: Supervised Instruction Tuning Meets Question Answering for Improved Dialogue NLU}

\author{~Evgeniia Razumovskaia$^\mathbf{1}$, ~Goran Glava\v{s}$^\mathbf{2}$, Anna Korhonen$^\mathbf{1}$, ~\textbf{Ivan Vuli\'{c}$^\mathbf{1,3}$}\\
  $^\mathbf{1}$ Language Technology Lab, University of Cambridge \\
  $^\mathbf{2}$ Center for Artificial Intelligence and Data Science, University of W\"{u}rzburg\\
 $^3$ PolyAI Limited \\
  \texttt{er563@cam.ac.uk goran.glavas@uni-wuerzburg.de 
\{alk23,iv250\}@cam.ac.uk}}

\begin{document}
\maketitle
\begin{abstract}
Task-oriented dialogue (\tod) systems help users execute well-defined tasks across a variety of domains (e.g., \textit{flight booking} or \textit{food ordering}), with their Natural Language Understanding (NLU) components being dedicated to the analysis of user utterances, predicting users' intents (\textit{Intent Detection}, ID) and extracting values for informational slots (\textit{Value Extraction}, VE). In most domains, labelled NLU data is scarce, making sample-efficient learning -- enabled with effective transfer paradigms -- paramount. In this work, we introduce \method, a new framework for dialog NLU based on (i) instruction tuning and (ii) question-answering-based formulation of ID and VE tasks. According to the evaluation on established NLU benchmarks, \method sets the new state of the art in dialogue NLU, substantially surpassing the performance of current models based on standard fine-tuning objectives in both in-domain training and cross-domain transfer, and it also surpasses off-the-shelf large language models for the same task, both in terms of performance and inference efficiency. 
Furthermore, \method yields particularly large performance gains in cross-domain transfer, owing to the fact that our QA-based instruction tuning leverages similarities between natural language descriptions of classes (i.e., slots and intents) across domains.
\end{abstract}

\section{Introduction}
Task-oriented dialogue (\tod) systems support users in execution of specific, well-defined tasks through natural language interaction (e.g., ordering food or purchasing tickets) \cite{young2002talking,budzianowski-etal-2018-multiwoz}. Fine-grained understanding of user's utterances, commonly referred to as (dialogue) natural language understanding (NLU) is necessary for successful \tod \cite{larson-etal-2019-clinc150,casanueva-etal-2022-nluplusplus}. NLU modules of \tod systems typically solve two complementary tasks: (1) \textit{Intent detection} (ID) aims to recognise the purpose (i.e., intent) of the user’s utterance, classifying utterances into a set of predefined classes (e.g., the intent \texttt{lost\_luggage} in \textit{flight booking}); (2) \textit{Value extraction} (VE) aims to extract spans that express values for any of the predefined informational slots (e.g., a dialog system for \textit{booking flights} would have slots such as \texttt{origin}, \texttt{destination}, \texttt{time}, \texttt{maximal\_price}).  
Realistic \tod setups for both ID and VE typically involve a relatively large number of labels (e.g., >100 different intent classes), commonly with a limited number of labelled instances per class. Successfully addressing these tasks thus amounts to enabling sample-efficient learning by means of transferring knowledge from other tasks \cite{gao-etal-2019-dialog}, languages \cite{hung-etal-2022-multi2woz,moghe2022multi3nlu++}, or domains \cite{hung2022ds,moghe2022multi3nlu++}.   


In recent years -- in line with general NLP trends -- most NLU models \cite[inter alia]{budzianowski2019hellogpt2,hosseini2020simple,henderson-vulic-2021-convex} were obtained via standard, task-specific fine-tuning of pretrained Transformer-based language models (PLMs) \cite{devlin-etal-2019-bert,radford2019language}. Standard fine-tuning comes with task-specific (discriminative) objectives -- different from LM-ing as the pretraining objective -- which in principle impedes both knowledge transfer (1) from pretraining to downstream tasks and (2) between different downstream tasks. 
Prompting in contrast \cite{liu2023pretrainprompt} recasts downstream tasks into language modelling, making them more aligned with the models' pretraining. Finally, instruction-tuning \cite{sahn2022T0,chung2022flan} -- supervised training in which prompts created from instances are prepended with natural language descriptions of the tasks -- facilitate the transfer between arbitrary tasks, leveraging the generalisation over task descriptions for zero-shot inference (i.e., inference for tasks unseen in training). 
Despite the impressive zero-shot and in-context few-shot inference abilities of the more recent Large LMs (LLMs) \cite{brown2020gpt3,chowdhery2022palm,touvron2023llama}, supervised fine-tuning still brings substantial performance gains for dialog NLU \cite{hudevcek2023llms} and standard NLU tasks, even with low-resource supervision \cite{gao-etal-2021-making}.  

As generalisation to new domains (with limited in-domain annotation effort) is one of the main desiderata of \tod, some recent work on dialog NLU \cite{fuisz2022qasl,casanueva-etal-2022-nluplusplus} has recognised that ID and VE can be cast as question answering (QA) tasks: this facilitates transfer from models trained on large QA datasets \cite{rajpurkar-etal-2016-squad,lee-etal-2020-squad2}, allowing also to capitalise on other large datasets previously recast as QA \cite{mccann2018decathlonnlp,wang2022supernaturalinstructions}. These efforts, however, amount to sequential transfer with standard fine-tuning for QA and thus (i) do not align their fine-tuning with the models' pretraining objective; and without an LM-based objective they (ii) cannot benefit from cross-task transfer via natural language task formulations.    

\rparagraph{Contributions} Motivated by the above observations, we propose a new framework for dialogue NLU driven by QA-based instruction tuning. In \textbf{\method} (Supervised Question Answering Tuning on INstructions for dialogue NLU), we reformulate ID and VE into QA-based natural language instructions and, starting from a massively instruction-tuned PLM \cite{chung2022flan}, fine-tune it for our tasks relying on a small number of in-domain examples. The rationale behind \method is two-pronged: (1) transfer with a model that was previously instruction-tuned at scale improves the efficiency of learning from task-specific samples -- this is highly desirable in most \tod domains, where one typically deals with only a handful of labelled utterances; (2) while small-scale ID/VE instruction-tuning specialises the model for a particular \tod domain (e.g., \textit{restaurant booking}), the negligible size of in-domain training (compared to model's massive instruction-``pretraining'') should prevent overfitting to the \tod training domain and allow for effective cross-domain transfer.  

Our results strongly support both of the above assumptions: \method yields state-of-the-art performance on two prominent dialogue NLU benchmarks both in in-domain and cross-domain evaluations. 
%
\method brings particularly large gains in transfer between close \tod domains: classes in these domains have similar prompt descriptions, unlike the existing approaches based on standard fine-tuning. The code is openly available at \url{https://github.com/cambridgeltl/sqatin/}.


\section{\method: Methodology}
\label{sec:methodology}


\sparagraph{Standard Classification vs. Instruction Tuning for Dialog NLU} 
ID and VE are two tasks that comprise most Dialogue NLU modules. 
%
Both tasks are commonly cast as classification tasks: ID as a sequence classification task (i.e., one or more intent labels assigned for the whole utterance) and VE as a span extraction task, i.e., token-level classification.  

In standard classification with pretrained LMs, a task-specific classifier $c_t: \mathbf{X} \in \mathbb{R}^h \mapsto \mathcal{P}(C_t)$ converts $h$-dimensional sequence or token representations (output by the LM) into a multinomial probability distribution over the set of task classes $C_t$. 
This means that a classifier $c_t$, trained for task $t$ with classes $C_t$, cannot be used to make predictions for any other classification task $t'$ with a different set of classes $C_{t'}$: thus, transfer between tasks can only occur indirectly through the parameters of the LM. This is particularly unfortunate for domain transfer in dialog NLU, where different domains often have semantically overlapping ID and VE classes (e.g., intent \texttt{confirm\_order} is essentially the same intent in \textit{flight booking} and in \textit{food ordering}).     
In contrast, instruction-tuning recasts classification as a language modelling (i.e., generation) task $LM: \mathbf{x} \in \mathbb{R}^h \mapsto \mathcal{P}(V_t)$, with $V_t$ as the subset of the LM's vocabulary where each token $v_t \in V_t$ represents one class $c_t$. This removes the need for a task-specific classifier (on top of the LM) and facilitates transfer between tasks, especially those with semantically overlapping class tokens.      

\rparagraph{QA-Based Instruction Tuning in \method} 
For the above reasons, we adopt an instruction tuning approach to ID and VE. We start from models that have been instruction-tuned at scale~\cite{wang2022instruction1600tasks,chung2022flan}, since these models 
come with a strong inductive bias to complete any new task expressed as an instruction, exhibiting impressive generalisation abilities (i.e., good performance on new tasks).  

\begin{figure}[!t]
    \centering
    \includegraphics[width=0.47\textwidth]{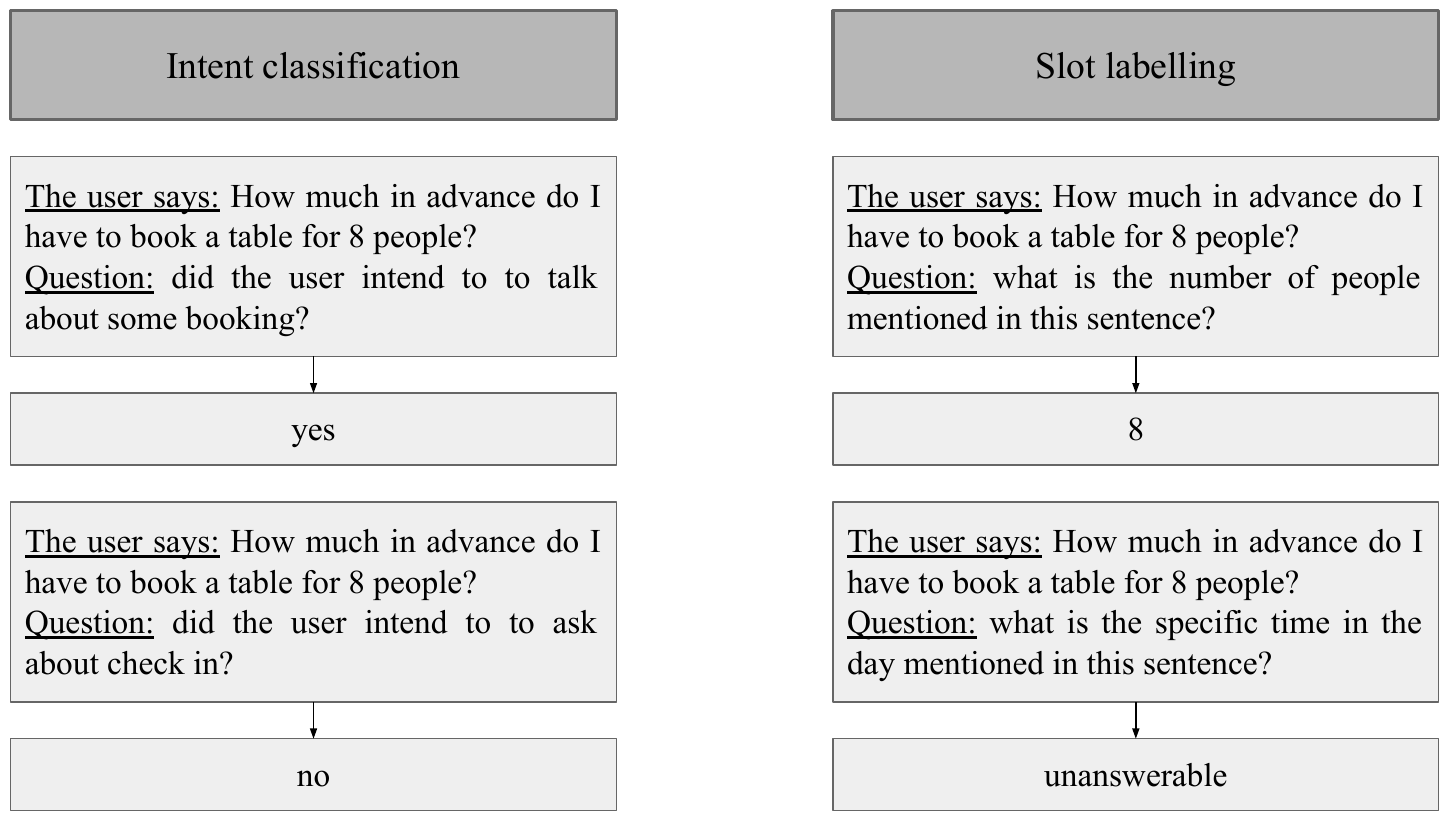}
    \caption{Instruction examples for ID and VE: for each we show one example where the class matches the utterance (i.e., for ID: correct intent class; for VE: a value for the slot class present) and one where it does not.}
    \label{fig:examples}
\end{figure}

As illustrated in Figure~\ref{fig:examples}, we formulate both ID and VE as text-to-text tasks, with our instruction input consisting of (i) \textit{context}, (ii) \textit{instance}, and (iii) \textit{prompt}. \textit{Context} (e.g., \textit{``The user says:''}) is the additional natural language description that is added (in our case, prepended) to the \textit{instance}, a user's utterance; \textit{Prompt} is the text that follows the \textit{instance} and describes the actual task, that is, what is to be predicted from the instance. 
We formulate \textit{prompts} as \textit{questions} for both tasks. The motivation for this is the fact that the instruction-tuned model from which we start \cite{chung2022flan} has been pretrained on QA formulations of various tasks and thus comes with an inductive bias for answering questions. For each training utterance, we create one instruction-based training example for each of the intent and slot classes: (1) for ID, the question incorporates a natural language description of the intent class (e.g., \textit{did the user \underline{intend to talk about some booking}? corresponds to the intent class \texttt{booking}}) and requires a binary answer (\texttt{yes} or \texttt{no}); (2) for VE, the question incorporates a natural language description of an informational slot (e.g., \textit{what is the \underline{number of people} mentioned?} corresponds to the slot \texttt{num\_guests}) -- the expected answer is the value for that slot, as expressed in the instance or \texttt{unanswerable} if the instance does not contain a value for the slot. 

A possible alternative to this ``one instruction per instance and class'' approach would be the more common prompt-based classification approach in which we create only one instruction per instance (e.g., with the question prompt \textit{``what is the intent of this sentence?''}) and the model is expected to generate the token of the correct intent, choosing between tokens of all intent classes. This, however, comes with two major drawbacks: (i) ID tasks commonly come with a large number of classes (e.g., more than 50) -- incorporating descriptions of all intent classes into a single prompt might thus surpass the input size of most models or they might struggle with memorizing all the options~\cite{Liu:2023arxiv}; (ii) ID is, in principle, a multi-label, rather than multi-class problem, which means that utterances can express more than just one intent -- this would require the model to output the text that somehow combines the tokens of more than one class, which is not something that instruction-based models have been pretrained for.

\begin{figure}[!t]
    \centering
    \includegraphics[width=0.48\textwidth]{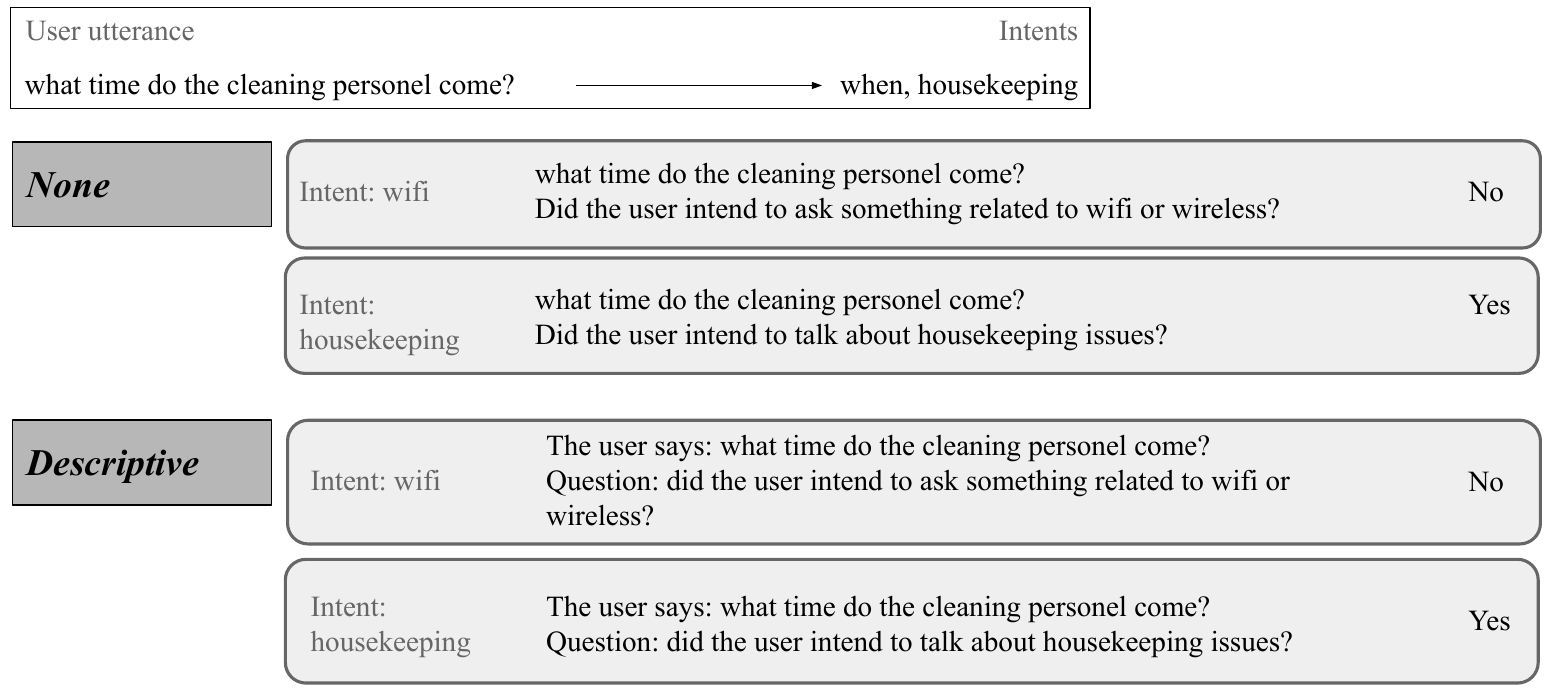}
    \caption{An annotated utterance from NLU++ transformed into corresponding \method instruction instances. For brevity, we display the transformation for only two intents (\texttt{wifi} and \texttt{housekeeping}), but the same transformation was applied for all intents.}\label{fig:instruction}
\end{figure}

We experimented with two different instruction formulations: (1) without context (\textit{None}), in which the instruction consists only of the instance and prompt; and (2) with descriptive context (\textit{Desc.}, where we prepend the utterance with \textit{``The user says:''} and the question prompt with \textit{``Question:''}, as illustrated Figure \ref{fig:instruction}. We selected these two particular instruction formulations (\textit{None} and \textit{Desc.}) based on their performance in a pilot study, which we describe in detail in the Appendix (\ref{app: different instructions}).



\section{Experimental Setup}\label{section: experimental setup}
We rely on the Flan-T5 instruction-pretrained models \cite{chung2022flan}. Unless stated otherwise, the main model is the \texttt{Base} variant. Training hyperparameters are described in detail in Appendix~\ref{app:hyperparameters}.


\rparagraph{Dialogue NLU Datasets} We run our experiments on two prominent dialogue NLU benchmarks: NLU++ \cite{casanueva-etal-2022-nluplusplus} and CLINC-150 \cite{larson-etal-2019-clinc150}. 
NLU++ contains user utterances from real conversations in two domains: \textit{banking} and \textit{hotels}.
NLU++ differs from most other \tod datasets in two important aspects: (i) it encompasses both generic (i.e., domain-universal) intents (e.g., \texttt{booking}) and slots (e.g., \texttt{date}) as well as the domain-specific ones (e.g., intent \texttt{credit\_card} in the \textit{banking} domain or slot \texttt{no\_rooms} in the \textit{hotels} domain) and (ii) its intents are ``factorized'' into ``atomic'' labels, with utterances then being assigned multiple intents (e.g., an utterance \textit{``wanna change my room reservation''} is labelled with three atomic intents -- \textit{change}, \textit{room}, and \textit{booking} -- rather than one complex intent \texttt{change\_room\_booking}).    
CLINC-150 encompasses over 20K utterances from 10 versatile domains (e.g., \textit{travel}, \textit{small talk}). Each domain has 15 intent labels, resulting in 150 intents in total. CLINC also contains utterances that do not belong to any of the 150 intents (labelled as \texttt{out\_of\_scope}).    
The fact that all CLINC domains have 15 intents, with the same number of instances per intent, allows for direct performance comparison across domains.\footnote{Prior work has mostly used CLINC-150 as a single-domain dataset with 150 intents, rather than multi-domain with domain-specific intents. 
In contrast, we are interested in cross-domain dialogue NLU performance and thus split the examples by domains. To ensure the replicability of results, we will make public the exact dataset splits that we used.}
%
With few-shot fine-tuning in focus, we evaluate the models in a folded cross-validation setup. NLU++ already comes with predefined splits for 10-fold and 20-fold cross-validation.\footnote{In the 20-fold setup, one fold contains $\approx 100$ utterances in the \textit{banking} domain and $\approx 50$ in the \textit{hotels} domain.} 
Analogously, we split data from each CLINC domain in 10 folds, resulting in 150 training examples per fold.

\rparagraph{Baselines} We compare \method against two types of state-of-the-art models for dialogue NLU. For brevity, we provide training and model selection details for both baselines in the appendix.

\rrparagraph{Classification from Sentence Embeddings (CL-SE)} Recent work on ID \cite{gerz2021mindsdataset, casanueva-etal-2022-nluplusplus} resorts to classifying -- with a shallow feed-forward classifier -- fixed sentence embeddings produced by of-the-shelf sentence encoders (SE). This avoids expensive fine-tuning of base LMs (e.g., RoBERTa) and yields comparable (or better) performance. We use LaBSE \cite{feng2022labse} as a state-of-the-art (SotA) SE.

\rrparagraph{Standard QA Fine-Tuning (QA-FT)} Similar to us, these models adopt a QA-based formulation of dialogue NLU but exclude the instruction component \cite{namazifar2021qanlu-amazon,casanueva-etal-2022-nluplusplus,fuisz2022qasl}. The key aspect is that the QA-based fine-tuning for ID and VE starts from the model that has previously been fine-tuned on large-scale QA datasets (e.g., SQUAD, \citet{rajpurkar2016squad,rajpurkar2018squad2.0}). To maximise comparability (given that \method is based on Flan-T5), we obtain our QA-FT baseline by fine-tuning the T5 model \cite{raffel2020t5model} previously trained on SQUAD 2.0.\footnote{We use the checkpoint at \url{https://huggingface.co/mrm8488/t5-base-finetuned-squadv2}.}
   


We report the standard micro-F1 scores. VE predictions are considered correct only if they exactly match the gold value span.  


\section{Main Evaluation}\label{sec: results}
\sparagraph{Preliminary Study: Zero-Shot ID \& VE} The key hypothesis behind \method is that instruction-tuned models have stronger inductive bias for dialogue NLU than models fine-tuned in the standard manner, including those trained for QA \citep{namazifar2021qanlu-amazon,fuisz2022qasl}. We thus preliminarily compare zero-shot ID/VE performance of (1) the instruction-trained Flan-T5 and (2) T5 fine-tuned for QA on SQUAD2.0 (denoted QA-T5) on NLU++. 
The results in Table \ref{tab:zero shot} show that Flan-T5 is much more robust ``out of the box''. While QA-T5 has better VE performance in the \textit{banking} domain, it yields near-zero performance in all other setups. This validates our selection of the instruction-tuned Flan-T5 as the starting point for \method.  


\begin{table}[!t]
\centering
{\fontsize{7.3}{7.5}\selectfont
\def\arraystretch{0.999}
\begin{tabularx}{0.999\linewidth}{l YYYY}
\toprule
\multirow{2}{*}{\bf Model} & \multicolumn{2}{c}{\bf ID} & \multicolumn{2}{c}{\bf VE} \\ \cmidrule(lr){2-3} \cmidrule(lr){4-5}
                       & \textit{20-Fold}   & \textit{10-Fold}   & \textit{20-Fold}    & \textit{10-Fold}   \\ \midrule
\rowcolor{Gray} \multicolumn{5}{c}{\bf \textsc{banking}}                                              \\ \midrule
QA-T5                     &    ~0.6        &     ~0.6      &      12.5      &    12.5       \\
Flan-T5                   &       21.9     &      21.9     &       ~3.2     &     ~3.2      \\ \midrule
\rowcolor{Gray} \multicolumn{5}{c}{\bf \textsc{hotels}}                                               \\ \midrule
QA-T5                     &    ~0.4        &     ~0.4      &      0.0      &  0.0         \\
Flan-T5                   &     20.9       &     21.9      &     5.9       &     5.8      \\ \bottomrule
\end{tabularx}
}
\caption{Zero-shot results for ID and VE on NLU++.}\label{tab:zero shot}
\end{table}


\begin{table}[!t]
\def\arraystretch{0.999}
{\fontsize{7.3}{7.5}\selectfont
\begin{tabularx}{0.99\linewidth}{ll YYYY}
\toprule
\multirow{2}{*}{\bf Model} & \multirow{2}{*}{\em Templ.} & \multicolumn{2}{c}{\bf ID}                                    & \multicolumn{2}{c}{\bf VE}                                    \\ \cmidrule(lr){3-4} \cmidrule(lr){5-6}
                       &                           & \textit{20-F} & \textit{10-F} & \textit{20-F} & \textit{10-F} \\ \midrule
\rowcolor{Gray} \multicolumn{6}{c}{\bf \textsc{banking}}                                                                                                                                                \\ \midrule
CL-SE &               &   58.1                       & 68.8                        & N/A    & N/A    \\
QA-FT: RoBERTa  &               &                 80.3         &     85.6                    & 50.5    & 56.7    \\
QA-FT: mDeBERTa           &                           & 80.8                       & 85.0                          & 59.7                       & 66.5                       \\ 
QA-FT: T5                 &                           & 82.7                       & 86.8                       & 61.5                       & 73.5                       \\ \midrule
\multirow{2}{*}{\method}  & \textit{None}                      & 85.6                       & \textbf{88.5}                       & 64.9                       & 75.4                       \\
                       & \textit{Desc.}                     & \textbf{85.8}                       & 88.4                       & \textbf{66.3}                       & \textbf{76.3}                       \\ \midrule
\rowcolor{Gray} \multicolumn{6}{c}{\bf \textsc{hotels}}                                                                                                                                                 \\ \midrule
CL-SE              &                           & 51.9                        & 61.8                        & N/A   & N/A    \\
QA-FT: RoBERTa               &                           &        67.4                 &              73.3           & 48.1    & 52.4    \\
QA-FT: mDeBERTa           &                           & 66.9                        & 73.2                        & \textbf{61.6}                        & 67.3    \\  
QA-FT: T5                 &                           & 69.2                       & 76.5                        & 57.2                       & \textbf{67.9}                       \\ \midrule
\multirow{2}{*}{\method}  & \textit{None}                      & 73.1                       & 78.0                       & 58.0                       & \textbf{67.7}                       \\
                       & \textit{Desc.}                     & \textbf{73.4}                       & \textbf{78.1}                       & 58.7                       & 67.0                       \\ \bottomrule
\end{tabularx}
}
\caption{In-domain ID and VE performance for \method and SotA baselines (CL-SE and QA-FT with different base models). \textbf{Bold:} best column score.
}
\label{tab: nlu plus plus in domain}
\end{table}


\begin{table*}[!t]
\def\arraystretch{0.999}
{\fontsize{7.3}{7.5}\selectfont
\begin{tabularx}{0.99\linewidth}{ll YYYYYYYYYYYY}
\toprule
\rowcolor{Gray} Model                  & Template & \textsc{auto}      & \textsc{banking}   & \begin{tabular}[c]{@{}c@{}}\textsc{credit}\\\textsc{card}\end{tabular} & \textsc{home} & \begin{tabular}[c]{@{}c@{}}\textsc{kitchen}\\\textsc{\&dining}\end{tabular} & \textsc{meta} & \begin{tabular}[c]{@{}c@{}}\textsc{small}\\\textsc{talk}\end{tabular} & \textsc{travel} & \textsc{utility} & \textsc{work} & \textsc{\textbf{AVG}} \\ \midrule
\multicolumn{2}{l}{CL-SE}    & 92.74                         & 92.30                         & 90.48                           & 88.58                    & 91.19                       & 90.19                    & 90.90                           & 95.29                      & 94.53                       & 91.93           & 91.81         
\\ \multicolumn{2}{l}{QA-FT: T5}        & 90.42                         & 94.38                         & 94.42                           & 89.23                    & 93.22                       & 90.10                     & 81.36                          & 97.67                      & 94.66                       & 89.99          & 91.54          \\ \midrule
                       & \textit{None}     & \textbf{94.47} & 96.04 & 95.64                           & 91.92                    & 95.01                       & 90.55                    & 93.10                          & \textbf{97.77}                      & 95.72                       & 91.56       & 94.18             \\
\multirow{-2}{*}{\method} & \textit{Desc.}    & \textbf{94.47}                         & \textbf{96.11}                         & \textbf{95.85}                           & \textbf{92.66}                    & \textbf{95.36}                       & \textbf{91.52}                    & \textbf{93.12}                          & 96.97                      & \textbf{96.07}                       & \textbf{92.01}  & \textbf{94.42}                  \\ \bottomrule
\end{tabularx}
}
\caption{In-domain ID results on CLINC-150 for \method and the baselines (CL-SE and QA-FT).
} \label{tab: clinc150 in domain}
\end{table*}

\begin{table}[!t]
\def\arraystretch{0.999}
{\fontsize{7.1}{7.3}\selectfont
\begin{tabularx}{0.99\linewidth}{ll YYYY}
\toprule
\multirow{2}{*}{\bf Model} & \multirow{2}{*}{\em Templ.} & \multicolumn{2}{c}{\bf ID}                                    & \multicolumn{2}{c}{\bf VE}                                    \\ \cmidrule(lr){3-4} \cmidrule(lr){5-6}
                       &                           & \textit{20-F} & \textit{10-F} & \textit{20-F} & \textit{10-F} \\ \midrule
\rowcolor{Gray} \multicolumn{6}{c}{\bf \textsc{banking $\rightarrow$ hotels}}                                                                                                                                                \\ \midrule
QA-FT: T5                 &                       &      66.70                 &       \textbf{69.68}                 &                  30.86          &           38.09             \\
\multirow{2}{*}{\method}  & \textit{None}                      &         66.68               &         68.18             &           \textbf{33.24}           &       \textbf{39.48}                 \\
                       & \textit{Desc.}                     &               \textbf{67.04}        &   68.48                     &     \textbf{33.24}                   &          37.41              \\ \midrule
\rowcolor{Gray} \multicolumn{6}{c}{\bf \textsc{hotels $\rightarrow$ banking}}                                                                                                                                                \\ \midrule
QA-FT: T5                 &                &          59.76             &        66.12                 &       35.08                &       44.60                 \\
\multirow{2}{*}{\method}  & \textit{None}                      &               65.35         &        67.34                &      44.72                 &   \textbf{52.05}  \\
                       & \textit{Desc.}                     &        \textbf{66.44}               &         \textbf{68.56}               &        \textbf{45.69}                &       51.87                \\ \bottomrule
\end{tabularx}
}
\caption{Domain transfer results for \method and the QA-FT (T5) baseline on NLU++ (between \textsc{banking} and \textsc{hotels}). \textbf{Bold:} best score in each column.} \label{tab: nlu++ cross domain}
\end{table}

\rparagraph{In-Domain Results} We next compare the supervised in-domain performance (i.e., training and test instances from the same domain) of \method against the CL-SE and QA-FT baselines. Tables \ref{tab: nlu plus plus in domain} and \ref{tab: clinc150 in domain} display the results on NLU++ and CLINC-150, respectively. On NLU++, we additionally provide QA-FT results with two other base models, RoBERTa \cite{liu2019roberta} and mDeBERTa \cite{he2021debertav3}, copied directly from \cite{casanueva-etal-2022-nluplusplus} and \cite{moghe2022multi3nlu++}, respectively.   


\method consistently and considerably outperforms the baseline models, on both tasks and on both datasets.  
These results confirm that instruction-based models have stronger inductive biases than QA-fine-tuned models: these biases are propagated in task-specific instruction-based fine-tuning, resulting in SotA performance. 
The gains seem more pronounced in setups with less training data (i.e., 20-Fold in Table \ref{tab: nlu plus plus in domain}) rendering instruction-tuning more sample efficient than  (QA-based) fine-tuning. Overall, \method seems to work slightly better with descriptive context prompts added to the instruction (compare \textit{Desc.} vs. \textit{None}).    

\begin{figure*}[!t]
    \centering
    \includegraphics[width=0.98\textwidth]{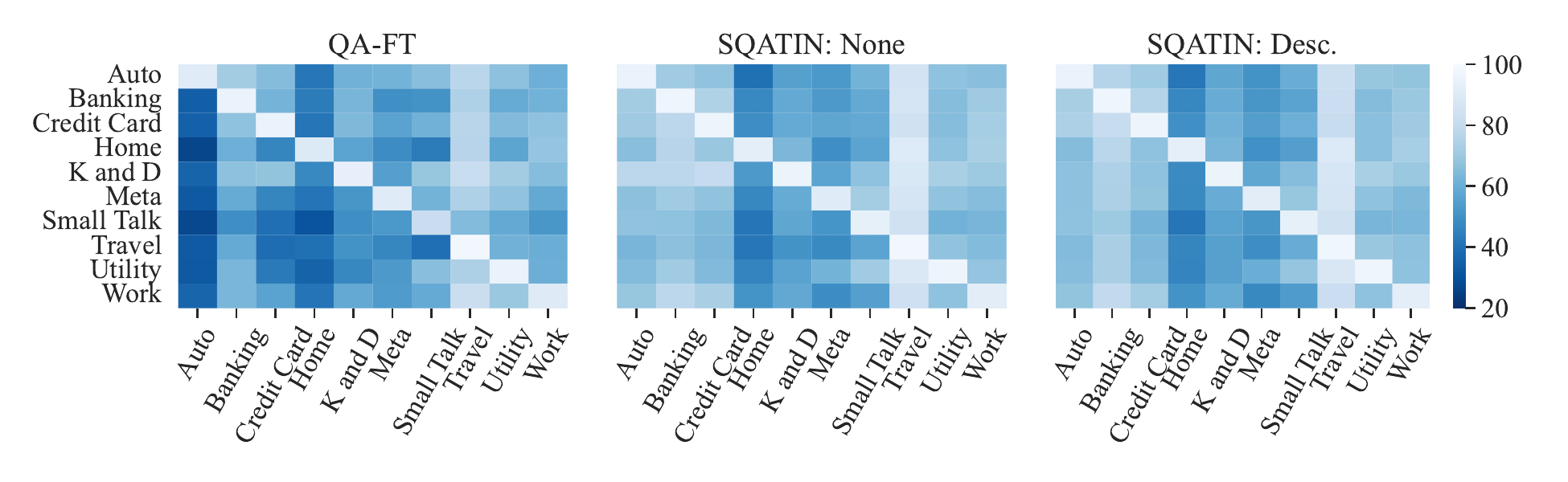}
    \vspace{-4mm}
    \caption{Cross-domain transfer results for ID on CLINC-150 for \method and the SotA QA-FT baseline. Full results in the tabular format are in Appendix~\ref{app: full cross-domain clinc results}. Diagonal values correspond to in-domain results. Source domains shown along the vertical axis and target domains along the horizontal axis.}\label{fig: clinc150 cross domain}
\end{figure*}

\rparagraph{Domain Transfer Results} We next train \method in one (source) domain and apply it in another (target) domain.
Table \ref{tab: nlu++ cross domain} and Figure \ref{fig: clinc150 cross domain} summarize the domain transfer results for NLU++ and CLINC-150 (all domain pairs), respectively.

Much like in in-domain training, \method consistently outperforms the SoTA baseline QA-FT in domain transfer (the only exception is \textsc{banking}$\rightarrow$\textsc{hotels} transfer for ID in the 10-Fold setup), only now by much wider margins for VE (e.g., by over 10 points in \textsc{hotels}$\rightarrow$\textsc{banking} transfer in the 20-Fold setup).
On CLINC-150, the results reveal not only that \method consistently outperforms QA-FT (consistently lighter heatmap cells for \method variants than for QA-T5) but that it is also able to better exploit label similarity between domains: e.g., for \textsc{credit card} as the target domain, \method obtains best performance when transferring from the \textsc{banking} domain, whereas QA-FT, in this case, finds \textsc{auto} as the best source.       



\rparagraph{Similarity of Intent Class Descriptions} 
Observing that \method yields best transfer performance between intuitively related domains, we now investigate more closely what type of similarity between domains drives the transfer: (i) similarity of examples (sim-E)  or (ii) similarity of intent class descriptions, incorporated in \method's prompts (sim-C). We quantify sim-E as the average similarity across all pairs of utterances between the domains: with  similarity of two utterances computed as cosine between their sentence embeddings, obtained with \texttt{mpnet} \cite{song2020mpnet} as the sentence encoder. Analogously, sim-C is computed as the average similarity of pairs of class prompts between the two domains.          %
%
%
We then measure the correlation (Pearson's $\rho$) between the transfer performance and sim-E or sim-C. Table \ref{tab: correlation intent description cross-domain} shows these correlations for each CLINC-150 domain as transfer target. Correlations are largest for domains that do have related domains in the dataset (e.g., \textsc{banking} and \textsc{credit card}) and lowest for domains that are quite different from all other (e.g., \textsc{auto} or \textsc{utility}). Importantly, sim-C shows higher average correlation with transfer performance than sim-E: this suggests that \method's instruction-based tuning with class descriptions in prompts truly captures similarities sets of intents and, consequently, especially improves transfer between related domains.    


\begin{table*}[!t]
\def\arraystretch{0.999}
{\fontsize{7.1}{7.5}\selectfont
\begin{tabularx}{0.99\linewidth}{l YYYYYYYYYYY}
\toprule
Template & \textsc{auto}                            & \textsc{banking} & \begin{tabular}[c]{@{}l@{}} \textsc{credit} \\ \textsc{card} \end{tabular} & \textsc{home} & \begin{tabular}[c]{@{}l@{}} \textsc{kitchen}\\ \textsc{\&dining} \end{tabular} & \textsc{meta} & \begin{tabular}[c]{@{}l@{}} \textsc{small}\\ \textsc{talk} \end{tabular} & \textsc{travel} & \textsc{utility} & \textsc{work} & \textsc{\textbf{avg}} \\ \midrule
\rowcolor{Gray}  \multicolumn{12}{c}{\bf In-Domain Training Examples} \\ \midrule
None     & -0.1443 & 0.5476                      & 0.4268                                                                     & 0.1318                   & 0.0204                      & 0.0970                   & 0.3279                                                                   & 0.0890                     & -0.2613                     & 0.5451                   & 0.2591                  \\
Desc.    & -0.1069 & 0.5710                      & 0.4695                                                                     & -0.1121                  & 0.1649                      & 0.0929                   & 0.1304                                                                   & -0.3360                    & -0.35                       & 0.6086                   & 0.2942                  \\ \midrule
\rowcolor{Gray}  \multicolumn{12}{c}{\bf Intent Descriptions}                                                                                                                                                                                                                                                                                                                                                                                                                \\ \midrule
None     & -0.2600     & 0.6260                      & 0.5076                                                                     & 0.3059                   & 0.1208                      & 0.2454                   & 0.6019                                                                   & 0.1633                     & 0.1388                      & 0.3830                   & 0.3353                  \\
Desc.    & -0.3376     & 0.5533                      & 0.5327                                                                     & 0.2319                   & -0.1091                     & 0.3165                   & 0.4884                                                                   & 0.1076                     & 0.0449                      & 0.4860                   & 0.3208                  \\ \bottomrule
\end{tabularx}
}
\caption{Correlation (Pearson's $\rho$) between domain transfer performance and domain similarity, measured in terms (i) of examples (sim-E) and (ii) class prompts (sim-C): shown for every CLINC-150 domain as the target.}\label{tab: correlation intent description cross-domain}
\end{table*}

\section{Further Analyses and Discussion}

\sparagraph{Cross-Task Generalisation}  
We next hypothesise that \method facilitates transfer between the two dialogue NLU tasks, given that \method's QA formulation conceptually allows for such cross-task transfer and presents both tasks to the model in the same format. 
Table~\ref{tab:cross task} compares the zero-shot ID performance of the off-the-shelf Flan-T5 (\textit{Non-tuned}) against the variant we \method-fine-tune for VE. We observe substantial improvements in ID after instruction-tuning for VE (around 5\% in the \textsc{banking} domain and over 10\% in the \textsc{hotels} domain), proving effective cross-task generalisation of \method in dialogue NLU. 
%

We then fine-tune the models \textit{jointl}y on ID and VE. Table~\ref{tab: task-specific vs joint} compares single-task training vs. multi-task training on both tasks. While multi-task training yields no clear gains for ID (as the easier of the two tasks), it gives consistent gains for VE (0.5-1.5 F1 points). This again indicates that \method facilitates transfer between the dialog NLU tasks. 



\begin{table}[!t]
\def\arraystretch{0.999}
{\fontsize{7.4}{7.6}\selectfont
\begin{tabularx}{0.999\linewidth}{l YYYY}
\toprule
 \multirow{2}{*}{{\bf Model}} & \multicolumn{2}{c}{\textsc{banking}} & \multicolumn{2}{c}{\textsc{hotels}} \\ \cmidrule(lr){2-3} \cmidrule(lr){4-5}
                      & \textit{20-Fold}    & \textit{10-Fold}   & \textit{20-Fold}    & \textit{10-Fold}   \\ \midrule
Non-tuned                     &     21.91       &     21.93     &  20.85     &     21.94     \\
Tuned for VE                   &        26.28         &   26.85        &   30.77         &    33.39      \\ \bottomrule
\end{tabularx}
}
\caption{\method's (\textit{Desc.} cross-task transfer performance on NLU++; VE$\rightarrow$ID.}\label{tab:cross task}
\end{table}

\setlength{\tabcolsep}{4.5pt}
\begin{table}[!t]
\def\arraystretch{0.999}
{\fontsize{7.4}{7.6}\selectfont
\begin{tabularx}{0.99\linewidth}{lll YYYY}
\toprule
\multirow{2}{*}{\bf Model} & \multirow{2}{*}{\em Template} &  & \multicolumn{2}{c}{\bf ID}                                    & \multicolumn{2}{c}{\bf VE}                                    \\ \cmidrule(lr){4-5} \cmidrule(lr){6-7}
                       &               &            & \textit{20-F} & \textit{10-F} & \textit{20-F} & \textit{10-F} \\ \midrule
\rowcolor{Gray} \multicolumn{7}{c}{\textsc{banking}}                                                                                                                                                \\ \midrule
\multirow{4}{*}{\method} & \multirow{2}{*}{\textit{None}}  & Single-task                     &        85.55             &        88.53            &       64.92              &        75.41             \\
                       &                       &  Multi-task                    &    85.69                  &           88.34          &    66.89               &  76.08                  \\
 & \multirow{2}{*}{\textit{Desc.}}  & Single-task                     &     85.78             &      88.41             &        66.32               &             76.26           \\
                       &                       &  Multi-task                    &      85.79                  &          88.42          &      67.88          &     76.76                 \\                       \midrule
\rowcolor{Gray} \multicolumn{7}{c}{\textsc{hotels}}                                                                                                                                                 \\ \midrule
\multirow{4}{*}{\method} & \multirow{2}{*}{\textit{None}}  & Single-task                     &       73.11              &        78.04             &             57.99           &         67.71               \\
                       &                       &  Multi-task                    &          72.70             &     77.73                 & 61.27                  &            68.66            \\
 & \multirow{2}{*}{\textit{Desc.}}  & Single-task                     &       73.35                 &            78.11            &          58.74              &             66.94           \\
                       &                       &  Multi-task                    &  73.15                     &               77.74         &   61.74                     &                 68.66       \\    \bottomrule
\end{tabularx}
}
\caption{Cross-task transfer: comparison between (in-domain) single-task (ID \textit{or} VE) and multi-task training (ID \textit{and} VE) on NLU++.}\label{tab: task-specific vs joint}
\end{table}

\rparagraph{Model Size}
 To analyse the impact of the underlying instruction-tuned model's size on performance, we also train \method on top of the following Flan-T5 models: \textsc{small} (80M parameters), \textsc{base} (250M) and \textsc{large} (780M), with the scores provided in Appendix~\ref{app: different model sizes hotels}. 
 \method yields strong in-domain performance even on top of the \textsc{small} Flan-T5. The margin between \textsc{large} and \textsc{base} is substantially smaller than that between \textsc{base} and \textsc{small}; for in-domain ID, the gap between \textsc{large} and \textsc{base} is negligible. The \textsc{small} models performs notably worse than its larger siblings only in cross-domain transfer, especially for VE. 
 Cross-domain performance of \textsc{large} almost reaches the in-domain performance of \textsc{small}, which is in line with observations that generalisation abilities of instruction-tuned models generally improve with their size \cite{chung2022flan}. 

\begin{figure}[!t]
    \centering
    \includegraphics[width=0.489\textwidth]{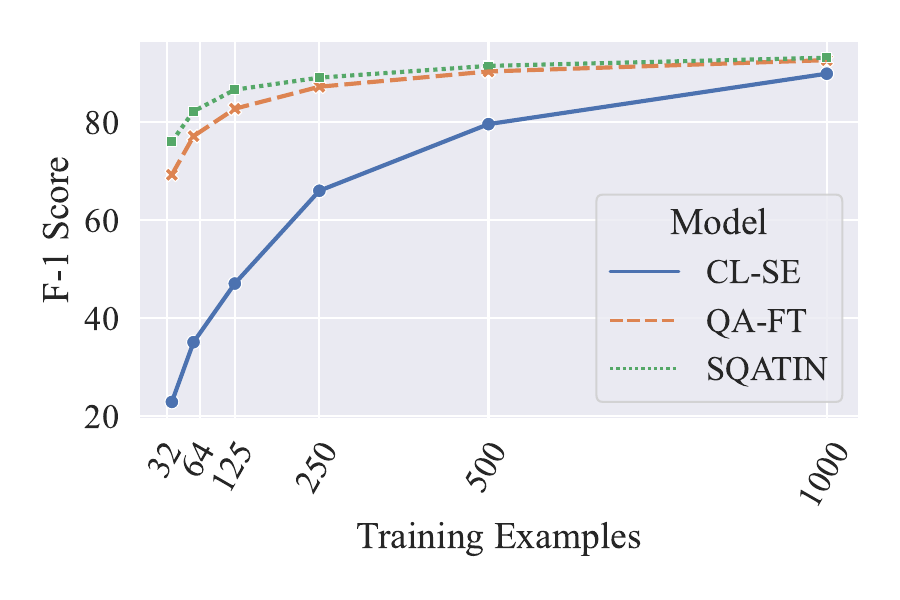}
    \vspace{-3.5mm}
    \caption{Comparison of ID models on \textsc{banking} domain on NLU++ for different training data sizes. The results are averages over 3 random seeds.}\label{fig: sample efficiency}
    \vspace{-2.5mm}
\end{figure}

\rparagraph{Sample Efficiency} Due to large-scale instruction pretraining, we expect \method to be more sample efficient than QA-FT and CL-SE. To test this, we train the models on training data of different sizes. The process is as follows: i) first, 1000 examples are randomly chosen for the test set; ii) from the rest we sample a random subset of $N$ training examples; 
iii) models are then trained on training set from step ii) and evaluated on test set from step i). This ensures that models trained on sets of different sizes are evaluated on the same test set, 
making the performances comparable. We use the same hyperparameter configuration from \S\ref{section: experimental setup} for all training sizes. Results in Figure~\ref{fig: sample efficiency} demonstrate that the scarcer the resources are, the more benefits \method brings over the baselines (QA-FT and especially CL-SE). Another observation is that both QA-based approaches, QA-FT as well as \method drastically outperform CL-SE in few-shot scenarios (cf. results for 32 and 64 training examples): this result justifies QA formulation for intent detection and value extraction in low-data setups.

\rparagraph{Base Instruction-Tuned Model} To evaluate the impact of the underlying instruction-tuned model, we apply SQATIN on top of Flan-T5-base and Tk-Instruct-base \cite{wang-etal-2022-tkinstruct}. The models are of the same size (250M), but pretrained on a different set of tasks (1,800+ tasks for Flan and 1,600+ tasks for Tk-Instruct). The experimental setup and training hyperparameters were kept the same across the models (see \S\ref{section: experimental setup} and Appendix \ref{app:hyperparameters} for the exact values). We run experiments on the intent detection task in the \textit{20-Fold} setup. The results presented in Table \ref{tab:flan_vs_tkinstruct} demonstrate improvements of SQATIN over QA-FT irrespective of the underlying instruction-tuned model. At the same time, the absolute scores do depend on the underlying model’s quality with Flan-T5 achieving better absolute scores.

\begin{table}[!t]
\def\arraystretch{0.999}
{\fontsize{7.4}{7.6}\selectfont
\begin{tabularx}{0.99\linewidth}{ll YYYY}
\toprule
\multirow{2}{*}{\bf Model} & \multirow{2}{*}{\em Model} & \multicolumn{2}{c}{\bf In-Domain}                                    & \multicolumn{2}{c}{\bf Cross-Domain}                                    \\ \cmidrule(lr){3-4} \cmidrule(lr){5-6}
                       &                           & \textsc{B} & \textsc{H} & \textsc{B $\rightarrow$ H} & \textsc{H $\rightarrow$ B} \\ \midrule
\multirow{2}{*}{QA-FT}  & N/A    & 82.7  &  69.2  & 66.7 &  59.8   \\
\cmidrule{2-6}

\multirow{2}{*}{\method} & \textit{Flan} & 85.8 & 73.4  & 67.4 & 66.4  \\
         & \textit{Tk-Instruct}  & 84.2 & 71.2 & 66.5  & 60.4\\     
                       \bottomrule
\end{tabularx}
}
\caption{Comparison of ID results in 20-Fold setup for different instruction-tuned models.}
\label{tab:flan_vs_tkinstruct}
\end{table}

\rparagraph{Independent QA versus Multiple-Choice} By design \method involves asking an independent question about every intent (for ID) and every slot (VE) from the ontology for each user utterance: this decomposition might impact inference efficiency. A more efficient alternative might be a common multiple-choice prompt-based approach, where we create one instruction per utterance and provide the model with all possible intent classes or slots. The model is then expected to generate all intents or slot values that apply to the given utterance in a single response. We use the same instruction formulations to ensure comparability and represent possible intent classes with natural language descriptions (e.g., ``to deny something'', ``to greet someone''); see an input example in Appendix~\ref{sec:input_example_mc}. Similarly to \method, we finetune an instruction-tuned model, namely, Flan-T5 (\textsc{base}), on the MC-style input. Training hyperparameters are provided in Appendix~\ref{app:hyperparameters}.

While offering potential benefits with inference speed, there are known deficiencies of this multiple-choice formulation (MC), as previously discussed in \S\ref{sec:methodology}. For instance, the average length (in tokens) of input of the independent, binary \method formualation for NLU++ ID and the MC formulation is 29.85 and 310.13, respectively. The difference might become even more salient with larger ontologies.
%
%
The results for NLU++ in Table \ref{tab:sqatin_vs_promptbased} demonstrate that the MC approach is considerably behind the independent-QA \method both in in-domain and cross-domain setups, regardless of the training data size or template formulation. This indicates that the per-intent or per-slot independent question formulation is necessary for sample-efficient generalisation of \method. We hypothesise that this is due to the data augmentation effects achieved this way. 

\rparagraph{\method versus In-Context Learning with ChatGPT} One alternative to supervised tuning of smaller models is in-context learning (ICL) with much larger instruction-tuned language models. ICL could be more computationally efficient at training time as it does not require fine-tuning the model while being more demanding at inference time, as the model size is considerably larger. To compare the performance of ICL with \method, we evaluate ChatGPT in two standard scenarios: (i) \textit{zero-shot (ZS)}, when the provided instruction includes task description with all possible options (intent descriptions in our case); and (ii) \textit{ICL}, when in addition to the above, the instruction also includes training examples which were used for supervised training in the models in every respective setting.\footnote{For the \textit{10-Fold} setup including all examples was impossible due to the context length limit. In this case, we fitted as many examples as possible by the context length.}  We evaluate \texttt{GPT-3.5-turbo-instruct} as the underlying model due to its strong ICL capabilities~\citep{ye2023gpt35turbocapabilities}.

Results in Table~\ref{tab:sqatin_vs_promptbased} demonstrate that \method performs consistently better than ChatGPT in both ZS and ICL scenarios. This suggests that even large models with ICL (and higher inference demands and cost) cannot surpass smaller highly specialised \method models for the fine-grained dialogue NLU tasks such as the NLU++ ones.


\begin{table}[!t]
\def\arraystretch{0.999}
{\fontsize{7.4}{7.6}\selectfont
\begin{tabularx}{0.999\linewidth}{ll YYYY}
\toprule
\multirow{2}{*}{\bf Model} & \multirow{2}{*}{\em Templ.} & \multicolumn{2}{c}{\bf In-Domain}                                    & \multicolumn{2}{c}{\bf Cross-Domain}                                    \\ \cmidrule(lr){3-4} \cmidrule(lr){5-6}
                       &                           & \textit{20-F} & \textit{10-F} & \textit{20-F} & \textit{10-F} \\ \midrule
\rowcolor{Gray} \multicolumn{6}{c}{\bf \textsc{banking}}                                                                                                                                                \\ \midrule
\multirow{2}{*}{ChatGPT ZS}  & N/A                      & 38.2                      & 38.2                       &         -- &         --       \\

\multirow{2}{*}{ChatGPT ICL}  & N/A                      & 67.5                      & 67.6                       &         -- &         --       \\ \cmidrule{2-6}

\multirow{2}{*}{\method}  & \textit{None}                      & 85.6                       & \textbf{88.5}                       &         66.7              &            68.2          \\
                       & \textit{Desc.}                     & \textbf{85.8}                       & 88.4                       &             \textbf{67.0}         &    \textbf{68.5}                \\ 
\multirow{2}{*}{MC}  & \textit{None}                      &       62.0                 &         67.9            &              39.3         &             46.1         \\
                       & \textit{Desc.}                     &          63.9             &    68.5              &            42.5          &          47.7          \\  \midrule
\rowcolor{Gray} \multicolumn{6}{c}{\bf \textsc{hotels}}                                                                                                                                                 \\ \midrule
\multirow{2}{*}{ChatGPT ZS}  & N/A                      & 39.1                      & 39.2                       &         -- &         --       \\

\multirow{2}{*}{ChatGPT ICL}  & N/A                      & 63.1                      & 67.9                       &         -- &         --       \\  \cmidrule{2-6}
\multirow{2}{*}{\method}  & \textit{None}                      & 73.1                       & 78.0                       &        65.4          & 67.3                      \\
                       & \textit{Desc.}                     & \textbf{73.4}                       & \textbf{78.1}                       &        \textbf{66.4}             &         \textbf{68.6}     \\ 
\multirow{2}{*}{MC}  & \textit{None}                      &            45.5            &            58.2         &              37.3         &          50.8            \\
                       & \textit{Desc.}                     &          50.0            &      59.7             &     41.3                 &        51.9            \\                       
                       \bottomrule
\end{tabularx}
}
\caption{Standard \method versus prompt-based multiple-choice (MC) task formulation for in-domain and cross-domain setups (ID on NLU++). 
}
\label{tab:sqatin_vs_promptbased}
\end{table}

\begin{figure}[!t]
    \centering
    \includegraphics[width=0.489\textwidth]{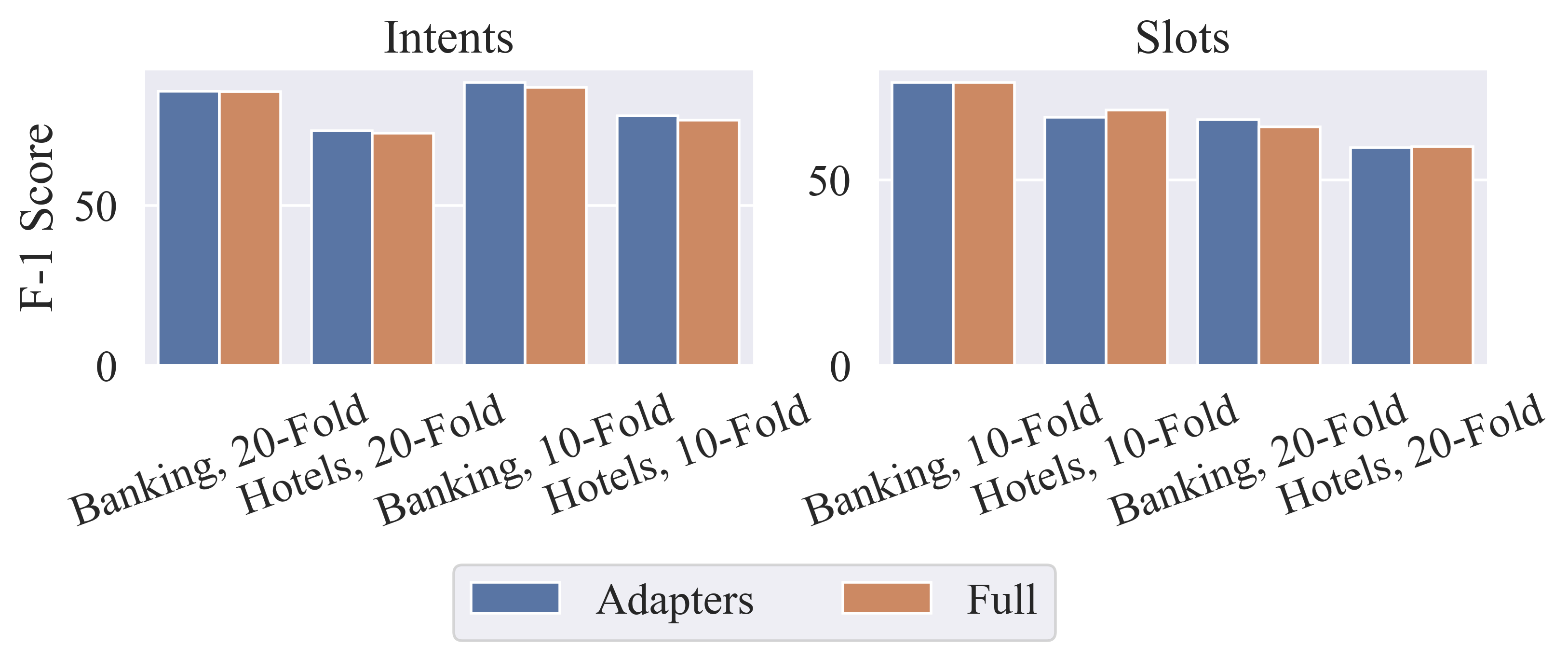}
    \caption{Full-model fine-tuning ($\approx$\,248M tunable parameters) versus PEFT with Adapters ($\approx$\,1.8M tunable parameters) in in-domain ID and VE. 
    }
    \label{fig: adapters results nlu++}
\end{figure}

\rparagraph{Parameter Efficiency} 
Next, we also investigate whether the performance benefits of \method extend when we replace full-model fine-tuning with the standard parameter-efficient fine-tuning (PEFT) methods \cite{ruder-etal-2022-modular} such as \textit{adapters}~\cite{houlsby2019parameter, pfeiffer2021adapterfusion}. In our case, relying on the standard bottleneck adapters with the reduction factor of 16~\cite{Poth:2023adapters}, for Flan-T5 \textsc{Base}, the number of tunable parameters is $\approx 250\times$ smaller than the size of the original model. The hyperparameters and training procedure are the same (see \S\ref{section: experimental setup}), except for the learning rate which was increased to $5e$-$4$.\footnote{Grid search over the set \{$5e$-$5$, $5e$-$4$, $5e$-$3$\} was run.} 
%
%
%
%
Figure~\ref{fig: adapters results nlu++} displays the performance of adapter-based fine-tuning on NLU++. The results render adapters extremely effective and comparable to full model fine-tuning, indicating that the benefits of \method are not tied only to full-model fine-tuning. 


\section{Related Work}
\sparagraph{Pretraining for \tod Dialogue} 
 %
 LLMs, trained on large web-scale corpora, revolutionised NLP, bringing massive performance gains to most NLP tasks. Besides general corpora, the most successful pretrained LMs for dialogue have have been additionally trained on more specialised, conversation-like data (e.g., from Reddit or Twitter). These models have been increasingly successful in both open-domain \cite[inter alia]{adiwardana2020meena, bao2020plato2, thoppilan2022lamda, dettmers2023qlora} and task-oriented dialogue \cite{budzianowski2019hellogpt2, lin-etal-2020-mintl, ham2020end, zhao2020learning}. Compared to general-purpose LM pretraining (e.g., BERT), dialogic pretraining has been shown to lead to higher performance in cross-domain transfer for dialogue NLU tasks \cite[interalia]{mi-etal-2021-self-training-bert-nlu,lin2021zeroshottodbert,hung2022ds} due to the versatility of texts used in pretraining. Another stand of work incestigated multi-task learning setups for dialogue NLU \cite{hosseini2020simple,liu2021pretrainingnoisychanelfortod, su2022multitaskplugandplaytod}. 
 In this work, in contrast, we resorted to models \textit{pretrained} on multiple tasks with instruction-based objectives, resulting with stronger inductive biases for cross-domain and cross-task settings. To the best of our knowledge, this work is the first to propose a unified (QA- and instruction-based) framework for both dialogue NLU tasks (ID and VE). 




\rparagraph{Instruction Tuning for Dialogue NLU} Instruction tuning is an emergent framework in NLP where a generative model completes a task by following natural language \textit{instructions}, possibly including few labelled instances following the instruction to make the whole prompt. These models generalise particularly well to tasks unseen during training \cite{chung2022flan, chowdhery2022palm} due to their ability to leverage the information about a task during inference \cite{liu2023pretrainprompt}. The performance, especially in zero-shot setup, is highly dependent on task definitions \cite{liu2023pretrainprompt} or providing several training examples \cite{min2022metaicl} in the instruction text (commonly known as in-context learning).\footnote{We note for completeness that prior work proposed to utilise masked language modelling abilities of pretrained models on standard, non-dialogue NLU tasks via cloze-style phrases \cite{schick-schutze-2021-epet} and demonstrated the potential of using few labelled examples for parameter updates \cite{schick-schutze-2021-pet20}. \citet{gao-etal-2021-making} show the effectiveness of prompt-based fine-tuning of masked language models when several task demonstrations are included into the input context. This is in contrast to \method which \textbf{a)} focuses on dialogue NLU tasks; \textbf{b)} utilises instruction-tuned generative language models and relies on their instruction-following, rather than mask-filling capabilities.} Dialogue follows the same trend: recent work \cite{gupta2022instructdial} demonstrated the zero-shot effectiveness of instruction-tuned models on dialogue tasks. Instruction engineering \cite{gupta2022instructdial, ruder2023xtremeup} and increasing the number of in-context instances can further improve the models' performance \cite{madotto2021few, mi2022cins}. The input (context) size of the models, however, puts a limit on the number of (1) training examples (2) classes (i.e., their descriptions) one can include in the prompt. \method deals with the issue in two ways: a) by recasting the dialogue NLU tasks as independent QA, at inference time we remove the need for the model to see all class descriptions at once; 
and b) we allow the model to learn from training examples in supervised fashion (versus in-context) thus not being limited by the base model's input length. We empirically validate that both have strong positive impact on task performance.



\section{Conclusion}
We have introduced a novel framework for dialogue NLU, \method, which combined (i) supervised instruction tuning and (ii) question-answering formulation of intent detection and value extraction. We evaluated \method on two established dialogue NLU benchmarks, demonstrating that \method brings substantial and consistent improvements over the existing SoTA approaches. The performance gains are especially pronounced in cross-domain transfer, as \method can leverage similarities between classes across domains via their descriptions. \method also performs well in cross-task transfer, enabling the two dialogue NLU tasks to benefit from one another. We also show that \method supports parameter-efficient fine-tuning and that it largely outperforms ICL with much larger (and more expensive) language models.


\section*{Limitations}

The majority of our experiments are based on the Flan collection of models as they were pretrained on a wide collection of tasks. However, we note that there are other instruction-based models \cite[inter alia]{ouyang2022instructgpt3, sahn2022T0, zhang2022optmodel}, with more getting published almost on a daily basis, which could be used with the proposed method and the choice of the instruction-based model is orthogonal to the proposed methodology. We leave wider exploration in this direction as future work. 

Additionally, we have focused on a single-source transfer across domains, i.e., a model trained on one domain was expected to be able to transfer to a multitude of others.  Future work will also explore the multi-source cross-domain transfer where the model would be finetuned on combined data from several domains and tested on data from domains not included in training.

In the evaluation, we rely on available standard dialogue NLU benchmarks built specifically to test few-shot in-domain and cross-domain generalisation abilities of the models. It is important to note that the benchmarks are only for English dialogue NLU.  We opt to confirm the effectiveness of \method in multilingual settings in future work. Exploration of \method in multilingual settings would be also dependent on the availability of strong multilingually pretrained instruction-based models. 

NLU++ includes the descriptions of intent and slot classes which means that they did not have to be created specifically for \method. It is important to note that when applying it to other datasets, such descriptions would need to be created manually, and there is no guarantee that the descriptions in NLU++ are optimal for fine-tuning.

Lastly, due to the computational cost of finetuning instruction-based models we largely rely on instruction wordings and training hyperparameters from prior work. We hope to perform a more detailed hyperparameter search in both wording of the instructions and training hyperparameters in the future, which might yield even higer absolute scores with SQATIN.

\section*{Acknowledgments}
We thank the anonymous reviewers for their helpful feedback. The work has been in part supported by a Huawei research donation to the Language Technology Lab at
the University of Cambridge. It has also been supported by the UK Research and Innovation (UKRI) Frontier Research Grant EP/Y031350/1 EQUATE (the UK government’s funding guarantee for ERC Advanced Grants) awarded to Anna Korhonen at the University of Cambridge. The work of Ivan Vuli\'{c} has been supported in part by a Royal Society University Research Fellowship \textit{‘Inclusive and Sustainable Language Technology for a Truly Multilingual World’} (no 221137; 2022-).

\bibliography{anthology,custom}

\clearpage
\appendix

\section{Different Instruction Formulations}\label{app: different instructions}

Choosing the right instruction formulation is often crucial (or at least important) to obtain strong performance from the instruction-based models. Thus, we conducted a pilot study for picking an optimal one. We experiment with 4 \textit{context} options, 4 options of text preceding a question and 3 \textit{prompt} options. The options (shown in Table \ref{app tab: instruction option}) were adapted from the templates used to train the Flan models \cite{chung2022flan}. We use Fold-0 of 10-Fold in-domain setting for intent detection to determine the best instruction formulation. 

The results of the preliminary study are shown in Table \ref{app tab: different instructions}. Although the range of results is not that large, we focus on two instruction formulations in further experiments: \texttt{none-none-none} and \texttt{usersaid-question-none}. The former is picked for similarity with the simple question answering formulation, although it leads to a lower performance.  This enables direct comparison to QA-based models. As this formulation contains only the input sentence and the questions (no description of the task or its context), we denote it as \textit{None}. The former instruction formulation (\texttt{usersaid-question-none}) is used as it contains the description of the context of the task and it led to the highest performance in the pilot study. As it contains a short description of the task, we denote it as \textit{Descriptive (Desc.)}.

\begin{table}[!t]
\centering
\resizebox{\linewidth}{!}{%
\begin{tabular}{@{}l@{}}
\toprule
\rowcolor{Gray} \textit{Context}      \\ \midrule
 \tabitem  ``''  [\texttt{none}]\\ 
 \tabitem  ``Given the following sentence: '' [\texttt{given}]\\
 \tabitem  ``Sentence: ''  [\texttt{sent}]\\
 \tabitem  ``The user says: ''  [\texttt{usersaid}]  \\ \midrule
\rowcolor{Gray} \textit{Pre-question} \\ \midrule
\tabitem  ``''  [\texttt{none}] \\ 
\tabitem  ``Question: ''  [\texttt{question}] \\ 
\tabitem  ``Based on the question: ''  [\texttt{based}] \\ 
\tabitem  ``Based on the question above: ''  [\texttt{basedabove}] \\ \midrule
\rowcolor{Gray} \textit{Prompt}       \\ \midrule
\tabitem  ``''  [\texttt{none}]\\
\tabitem  ``Answer: '' [\texttt{answer}] \\
\tabitem  ``Options: -yes -no\\ ~~~~~Answer:''           [\texttt{answeroptions}] \\ \bottomrule
\end{tabular}
}
\caption{Variants of instruction formulation.} \label{app tab: instruction option}
\vspace{-1.5em}
\end{table}

\begin{table}[!t]
{\footnotesize
\resizebox{\linewidth}{!}{%
\begin{tabular}{@{}lllrrr@{}}
\toprule
\rowcolor{Gray} Context & Pre-question & Prompt & Banking & Hotels & AVG \\ \midrule
none                                          & none                                          & none                                        & 77.2                                                            & 67.3                                                           & 72.25                            \\
sent                                          & none                                          & none                                        & 81.31                                                           & 76.45                                                          & 78.88                            \\
none                                          & none                                          & answer                                      & 80.96                                                           & 77.14                                                          & 79.05                            \\
given                                         & none                                          & none                                        & 81.4                                                            & 76.96                                                          & 79.18                            \\
none                                          & none                                          & answer-options                              & 81.22                                                           & 77.26                                                          & 79.24                            \\
none                                          & based-above                                   & answer                                      & 82.65                                                           & 75.9                                                           & 79.28                           \\
usersaid                                      & none                                          & none                                        & 81.72                                                           & 77.35                                                          & 79.54                           \\
given                                         & none                                          & answer                                      & 81.49                                                           & 77.69                                                          & 79.59                            \\
sent                                          & none                                          & answer                                      & 81.36                                                           & 78.06                                                          & 79.71                            \\
none                                          & based                                         & answer                                      & 82.1                                                            & 77.33                                                          & 79.72                           \\
none                                          & based                                         & answer-options                              & 82.1                                                            & 77.37                                                          & 79.74                           \\
sent                                          & based                                         & none                                        & 82.13                                                           & 77.38                                                          & 79.76                           \\
sent                                          & based-above                                   & none                                        & 82.68                                                           & 77                                                             & 79.84                            \\
sent                                          & based-above                                   & answer                                      & 82.73                                                           & 77.06                                                          & 79.90                           \\
sent                                          & based                                         & answer                                      & 82.15                                                           & 77.74                                                          & 79.95                           \\
none                                          & based-above                                   & answer-options                              & 82.67                                                           & 77.24                                                          & 79.96                           \\
sent                                          & none                                          & answer-options                              & 81.4                                                            & 78.63                                                          & 80.02                           \\
none                                          & based                                         & none                                        & 82.08                                                           & 78.1                                                           & 80.09                            \\
usersaid                                      & based                                         & none                                        & 82.34                                                           & 77.92                                                          & 80.13                            \\
usersaid                                      & none                                          & answer-options                              & 82.05                                                           & 78.28                                                          & 80.17                           \\
given                                         & none                                          & answer-options                              & 81.7                                                            & 78.63                                                          & 80.17                           \\
given                                         & question                                      & answer                                      & 83.49                                                           & 76.94                                                          & 80.22                           \\
sent                                          & based-above                                   & answer-options                              & 82.8                                                            & 77.65                                                          & 80.23                           \\
none                                          & based-above                                   & none                                        & 82.57                                                           & 77.93                                                          & 80.25                            \\
none                                          & question                                      & answer                                      & 83.17                                                           & 77.35                                                          & 80.26                            \\
sent                                          & question                                      & none                                        & 83.25                                                           & 77.27                                                          & 80.26                            \\
usersaid                                      & based                                         & answer                                      & 82.39                                                           & 78.15                                                          & 80.27                            \\
sent                                          & question                                      & answer                                      & 83.39                                                           & 77.29                                                          & 80.34                            \\
usersaid                                      & based                                         & none                                        & 82.99                                                           & 77.72                                                          & 80.36                           \\
usersaid                                      & based                                         & answer                                      & 83.05                                                           & 77.68                                                          & 80.37                           \\
none                                          & question                                      & answer-options                              & 83.22                                                           & 77.61                                                          & 80.42                           \\
given                                         & question                                      & answer-options                              & 83.6                                                            & 77.39                                                          & 80.50                           \\
usersaid                                      & none                                          & answer                                      & 81.83                                                           & 79.17                                                          & 80.5                             \\
sent                                          & based                                         & answer-options                              & 82.29                                                           & 78.78                                                          & 80.56                           \\
given                                         & question                                      & none                                        & 83.42                                                           & 77.66                                                          & 80.54                            \\
usersaid                                      & based                                         & answer-options                              & 82.42                                                           & 78.67                                                          & 80.55                           \\
sent                                          & question                                      & answer-options                              & 83.4                                                            & 77.7                                                           & 80.55                            \\
usersaid                                      & based                                         & answer-options                              & 83.08                                                           & 78.44                                                          & 80.76                            \\
none                                          & question                                      & none                                        & 83.08                                                           & 78.5                                                           & 80.79                            \\
usersaid                                      & question                                      & answer                                      & 83.88                                                           & 77.74                                                          & 80.81                            \\
usersaid                                      & question                                      & answer-options                              & 84.2                                                            & 77.43                                                          & 80.82                            \\
usersaid                                      & question                                      & none                                        & 83.85                                                           & 78.07                                                          & 80.96                            \\ \bottomrule
\end{tabular}
}
}
\caption{Performance of \method with different instruction wordings. The options are ordered in ascending average order.} \label{app tab: different instructions}
\end{table}

\section{Full Cross-Domain Results on CLINC-150 for Different Base Models}\label{app: full cross-domain clinc results}

The cross-domain results on CLINC-150 for QA-FT and different versions of \method are provided in Tables ~\ref{app tab: clinc150 cross domain qa baseline}, \ref{app tab: clinc150 cross domain our none baseline} and \ref{app tab: clinc150 cross domain our desc baseline}.

\begin{table*}[!t]
{\footnotesize
\resizebox{\linewidth}{!}{%
\begin{tabular}{lrrrrrrrrrr}
\toprule
\rowcolor{Gray}                                                       & \multicolumn{10}{c}{QA-FT pretrained on SQUAD 2.0}                                                                                                                                                                                                                                                                                                                                                                                                                                                                                                                \\ 
\rowcolor{Gray}                                                       & \textsc{auto} & \textsc{banking} & \begin{tabular}[c]{@{}l@{}}\textsc{credit} \\ \textsc{card}\end{tabular} & \textsc{home} & \textsc{k and d} & \textsc{meta} & \begin{tabular}[c]{@{}l@{}}\textsc{small}\\ \textsc{talk}\end{tabular} & \textsc{travel} & \textsc{utility} & \textsc{work} \\ \midrule
\textsc{auto}                                                   & 90.42                                            & 71.08                                               & 65.22                                                                                             & 42.03                                            & 61.23                                               & 61.78                                            & 65.64                                                  & 77.04                                              & 66.7                                                & 60.5                                             \\
\textsc{banking}                                                & 34.67                                            & 94.38                                               & 62.16                                                                                             & 43.35                                            & 62.51                                               & 49.43                                            & 50.35                                                  & 74.33                                              & 58.96                                               & 61.45                                            \\
\begin{tabular}[c]{@{}l@{}}\textsc{credit} \\ \textsc{card}\end{tabular} & 35.19                                            & 66.94                                               & 94.42                                                                                             & 41.28                                            & 64.05                                               & 55.86                                            & 61.13                                                  & 76.54                                              & 64.14                                               & 66.92                                            \\
\textsc{home}                                                   & 26.68                                            & 60.4                                                & 46.07                                                                                             & 89.23                                            & 55.95                                               & 48.64                                            & 43.35                                                  & 76.05                                              & 56.65                                               & 68.08                                            \\
\textsc{k and d}                                                & 35.96                                            & 66.85                                               & 67.75                                                                                             & 46.98                                            & 93.22                                               & 54.52                                            & 68.6                                                   & 80.95                                              & 71.08                                               & 65.5                                             \\
\textsc{meta}                                                   & 32.51                                            & 58.92                                               & 45.94                                                                                             & 41.11                                            & 51.25                                               & 90.1                                             & 61.68                                                  & 74.11                                              & 67.33                                               & 58.19                                            \\
\begin{tabular}[c]{@{}l@{}}\textsc{small} \\ \textsc{talk}\end{tabular}                                             & 27.2                                             & 49.17                                               & 39.61                                                                                             & 30.69                                            & 49.17                                               & 52.4                                             & 81.36                                                  & 64.59                                              & 58.16                                               & 51.62                                            \\
\textsc{travel}                                                 & 32.96                                            & 58.54                                               & 38.89                                                                                             & 39.71                                            & 50.6                                                & 46.53                                            & 39.46                                                  & 97.67                                              & 61.13                                               & 59.72                                            \\
\textsc{utility}                                                & 32.61                                            & 63.12                                               & 42.76                                                                                             & 35.91                                            & 46.87                                               & 52.67                                            & 65.77                                                  & 73.62                                              & 94.65                                               & 60.08                                            \\
\textsc{work}                                                   & 36.32                                            & 62.9                                                & 55.93                                                                                             & 41.05                                            & 58.24                                               & 53.14                                            & 58.62                                                  & 81.83                                              & 69.13                                               & 89.99                                            \\ \bottomrule
\end{tabular}
}
}
\caption{\textit{Cross-domain} intent detection using QA-based model on CLINC-150 \citep{larson-etal-2019-clinc150}. \textsc{K and D} stands for \textsc{Kitchen and Dining} domain. The rows are source domains while columns show target domains.} \label{app tab: clinc150 cross domain qa baseline}
\end{table*}

\begin{table*}[!t]
{\footnotesize
\resizebox{\linewidth}{!}{%
\begin{tabular}{lrrrrrrrrrr}
\toprule
\rowcolor{Gray}                                                       & \multicolumn{10}{c}{\method: \textit{None}}                                                                                                                                                                                                                                                                                                                                                                                                                                                                                                                \\ 
\rowcolor{Gray}                                                       & \textsc{auto} & \textsc{banking} & \begin{tabular}[c]{@{}l@{}}\textsc{credit} \\ \textsc{card}\end{tabular} & \textsc{home} & \textsc{k and d} & \textsc{meta} & \begin{tabular}[c]{@{}l@{}}\textsc{small}\\ \textsc{talk}\end{tabular} & \textsc{travel} & \textsc{utility} & \textsc{work} \\ \midrule
\textsc{auto}                                                   & 94.47                                            & 70.87                                               & 67.26                                                                                             & 39.75                                            & 54.96                                               & 52.2                                             & 61.57                                                  & 85.01                                              & 67.09                                               & 65.71                                            \\
\textsc{banking}                                                & 71.2                                             & 96.04                                               & 74.53                                                                                             & 46.92                                            & 58.31                                               & 52.81                                            & 58.3                                                   & 86.02                                              & 65.58                                               & 70.27                                            \\
\begin{tabular}[c]{@{}l@{}} \textsc{credit} \\ \textsc{card} \end{tabular} & 70.08                                            & 77.44                                               & 95.64                                                                                             & 48.97                                            & 58.71                                               & 57                                               & 58.4                                                   & 84.3                                               & 65.53                                               & 71.68                                            \\
\textsc{home}                                                   & 65.8                                             & 76.24                                               & 68.91                                                                                             & 91.91                                            & 63.3                                                & 49.18                                            & 56.1                                                   & 89.59                                              & 66.98                                               & 72.51                                            \\
\textsc{k and d}                                                & 77.25                                            & 77.38                                               & 79.84                                                                                             & 52.53                                            & 95.01                                               & 56.22                                            & 67.09                                                  & 88.01                                              & 72.75                                               & 69.7                                             \\
\textsc{meta}                                                   & 66.5                                             & 70.49                                               & 67.33                                                                                             & 46.85                                            & 59.05                                               & 90.55                                            & 71.51                                                  & 85.98                                              & 67.26                                               & 65.47                                            \\
\begin{tabular}[c]{@{}l@{}}\textsc{small} \\ \textsc{talk} \end{tabular}                                             & 67.36                                            & 67.07                                               & 63.8                                                                                              & 41.52                                            & 57.04                                               & 51.12                                            & 93.1                                                   & 83.94                                              & 61.43                                               & 62.68                                            \\
\textsc{travel}                                                 & 62.8                                             & 66.26                                               & 63.34                                                                                             & 41.94                                            & 50.58                                               & 47.71                                            & 55.97                                                  & 97.77                                              & 67.35                                               & 64.58                                            \\
\textsc{utility}                                                & 64.6                                             & 70.71                                               & 64.35                                                                                             & 45.68                                            & 55.88                                               & 61.6                                             & 70.91                                                  & 88.28                                              & 95.72                                               & 67.97                                            \\
\textsc{work}                                                   & 68.68                                            & 77.19                                               & 73.12                                                                                             & 50.89                                            & 58.03                                               & 48.63                                            & 54.5                                                   & 83.31                                              & 67.05                                               & 91.56                                            \\ \hline
\end{tabular}
}
}
\caption{\textit{Cross-domain} intent detection using \method on CLINC-150 \citep{larson-etal-2019-clinc150} with \textit{None} templates. \textsc{K and D} stands for \textsc{Kitchen and Dining} domain. The rows are source domains while columns show target domains.} \label{app tab: clinc150 cross domain our none baseline}
\end{table*}

\begin{table*}[!t]
{\footnotesize
\resizebox{\linewidth}{!}{%
\begin{tabular}{lrrrrrrrrrr}
\toprule
\rowcolor{Gray}                                                       & \multicolumn{10}{c}{\method: \textit{Desc.}}                                                                                                                                                                                                                                                                                                                                                                                                                                                                                                                \\ 
\rowcolor{Gray}                                                       & \textsc{auto} & \textsc{banking} & \begin{tabular}[c]{@{}l@{}}\textsc{credit} \\ \textsc{card}\end{tabular} & \textsc{home} & \textsc{k and d} & \textsc{meta} & \begin{tabular}[c]{@{}l@{}}\textsc{small}\\ \textsc{talk}\end{tabular} & \textsc{travel} & \textsc{utility} & \textsc{work} \\ \midrule
auto                                                   & 94.47                                            & 75.69                                               & 70.47                                                                                             & 41.68                                            & 56.88                                               & 50.47                                            & 59.61                                                  & 82.45                                              & 68.54                                               & 67.51                                            \\
banking                                                & 72.43                                            & 96.11                                               & 75.91                                                                                             & 46.77                                            & 59.13                                               & 51.44                                            & 55.68                                                  & 81.96                                              & 65.14                                               & 69.08                                            \\
\begin{tabular}[c]{@{}l@{}}credit \\ card\end{tabular} & 73.62                                            & 80.39                                               & 95.85                                                                                             & 49.55                                            & 61.13                                               & 54.34                                            & 60.59                                                  & 80.81                                              & 66.01                                               & 70.23                                            \\
home                                                   & 65.04                                            & 76.7                                                & 66.99                                                                                             & 92.66                                            & 62.81                                               & 49.83                                            & 54.21                                                  & 88.98                                              & 66.03                                               & 72.07                                            \\
k and d                                                & 66.79                                            & 73.88                                               & 66.92                                                                                             & 47.91                                            & 95.36                                               & 57.31                                            & 65.57                                                  & 87.18                                              & 72.71                                               & 69.37                                            \\
meta                                                   & 66.73                                            & 73.66                                               & 67.55                                                                                             & 47.56                                            & 59.12                                               & 91.52                                            & 68.59                                                  & 86.31                                              & 67.01                                               & 63.85                                            \\
\begin{tabular}[c]{@{}l@{}}small \\ talk\end{tabular}  & 67.08                                            & 69.89                                               & 61.95                                                                                             & 41.26                                            & 55.93                                               & 51.33                                            & 93.12                                                  & 84.28                                              & 62.62                                               & 62.97                                            \\
travel                                                 & 64.5                                             & 73.05                                               & 63.56                                                                                             & 46                                               & 54.73                                               & 48.81                                            & 59.14                                                  & 96.97                                              & 68.92                                               & 66.66                                            \\
utility                                                & 65.39                                            & 73.03                                               & 64.25                                                                                             & 45.66                                            & 55.26                                               & 59.82                                            & 68.29                                                  & 87.59                                              & 96.07                                               & 67.09                                            \\
work                                                   & 67.8                                             & 79.15                                               & 71.26                                                                                             & 50.41                                            & 58.86                                               & 47.48                                            & 53.41                                                  & 82.07                                              & 67.15                                               & 92.01                                            \\ \hline
\end{tabular}
}
}
\caption{\textit{Cross-domain} intent detection using \method on CLINC-150 \citep{larson-etal-2019-clinc150} with \textit{Descriptive} templates. \textsc{K and D} stands for \textsc{Kitchen and Dining} domain. The rows are source domains while columns show target domains.} \label{app tab: clinc150 cross domain our desc baseline}
\end{table*}

\begin{table}[!t]
{\footnotesize
\resizebox{\linewidth}{!}{%
\begin{tabular}{@{}lllcccc@{}}
\toprule
\multirow{2}{*}{Model} & \multirow{2}{*}{Template} \multirow{2}{*}{} & & \multicolumn{2}{c}{ID}                                    & \multicolumn{2}{c}{VE}                                    \\ \cmidrule(lr){4-5} \cmidrule(lr){6-7}
                       &                          &                           & 20-Fold & 10-Fold & 20-Fold & 10-Fold \\ \midrule
\rowcolor{Gray} \multicolumn{7}{c}{\textsc{banking} $\rightarrow$ \textsc{hotels}}                                                                                                                                                \\ \midrule
\multirow{4}{*}{\method} & \multirow{2}{*}{\textit{None}}  & Single-Task                     &       66.61       &       68.18    &      33.24     &     39.48          \\
                       &                       &  Multi-Task                    &   66.73           &  68.59             &      33.81  &     39.77          \\
 & \multirow{2}{*}{\textit{Desc.}}  & Single-Task                     &        67.04    &          68.48       &     33.25          &      37.41          \\
                       &                       &  Multi-Task                    &        67.28          &     68.15    &   33.08     &          36.18    \\                       \midrule
\rowcolor{Gray} \multicolumn{7}{c}{\textsc{hotels} $\rightarrow$  \textsc{banking}}                                                                                                                                                 \\ \midrule
\multirow{4}{*}{\method} & \multirow{2}{*}{\textit{None}}  & Single-Task                     &         65.35       &     67.34    &        44.72         &          52.05           \\
                       &                       &   Multi-Task                    &    64.68              &          67.06      &        45.38       &        51.44             \\
 & \multirow{2}{*}{\textit{Desc.}}  & Single-Task                     &     66.44                 &          68.56       &          45.69         &        51.87           \\
                       &                       &  Multi-Task                    &   66.86                  &        68.08              &         46.02          &    52.04                   \\    \bottomrule
\end{tabular}
}
}
\caption{Comparison of single-task and multi-task models for cross-domain intent detection and value extraction on NLU++.}\label{app tab:task-specific joint cross-domain}
\end{table}
\section{Comparison of Single-Task and Multi-Task Models for Cross-Domain Setups}

Comparison of cross-domain results of models trained with \method in single-task and multi-task setings is shown in Table \ref{app tab:task-specific joint cross-domain}.

\section{Fine-tuning and Hyperparameters}\label{app:hyperparameters}

The classifier of the CL-SE baseline is a feed-forward network with a single hidden layer of dimensionality 512 and \textit{tanh} as the non-linear activation function. With multi-label formulations of classification tasks (because instances in NLU++ can have multiple labels and those in CLINC-150 none), we apply \textit{sigmoid} as an output activation and train with the binary cross-entropy loss. 
At inference, we consider an intent class to be predicted if its probability, output of the sigmoid activation, is above the threshold $\theta = 0.3$.

The models are implemented using Transformers library \citep{wolf-etal-2020-transformers}. The models are loaded with sequence-to-sequence language modeling head. Baseline QA-based models and \method are fine-tuned with the same protocol and hyperparameters as in prior work \cite{casanueva-etal-2022-nluplusplus, fuisz2022qasl, moghe2022multi3nlu++}. They are trained for 10 epochs with the batch size of 8, with Adam optimizer \cite{kingma2014adamoptimizer} and the learning rate of 5e-5. Unless stated differently, we report the average cross-validation performance across all 10 or 20 folds the results are averages of 10 and 20 runs for 10- and 20-Fold setups, respectively.\footnote{We focus on the pre-defined few-shot 10-Fold and 20-Fold setups, as the baselines already demonstrate saturated performance on Large training data setups \cite{casanueva-etal-2022-nluplusplus}.}

\begin{figure}[t!]
    \centering
    \includegraphics[width=0.49\textwidth]{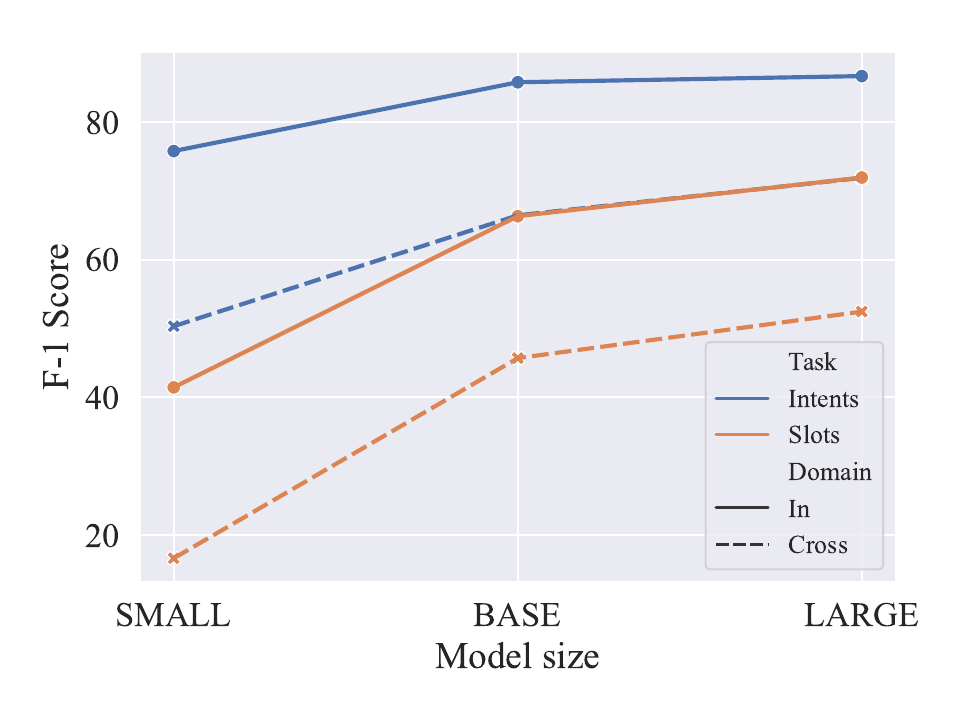}
    \caption{ID and VE performance (\textsc{banking} domain of NLU++, 20-Fold setup) for \method trained on top of Flan-T5 models of different sizes. Similar trends are observed in the \textsc{hotels} domain, see Figure~\ref{fig:modelsize hotels}. 
    }
    \vspace{-2mm}
    \label{fig:modelsize}
    \vspace{-1mm}
\end{figure}
\begin{figure}[!t]
    \centering
    \includegraphics[width=0.49\textwidth]{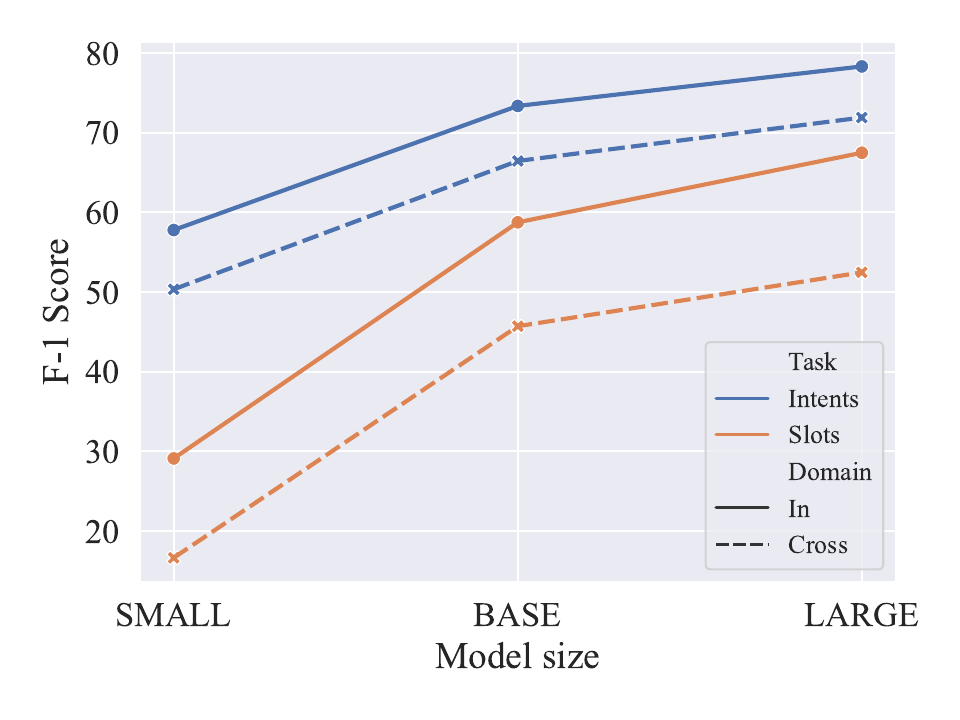}
    \caption{ID and VE performance (\textsc{hotels} domain of NLU++, 20-Fold setup) for \method trained on top of Flan-T5 models of different sizes. 
    }
    \label{fig:modelsize hotels}
\end{figure}
\section{Results for Different Model Sizes}
\label{app: different model sizes hotels}

The results for different model sizes for the two domains of NLU++ are plotted in Figure~\ref{fig:modelsize} and Figure~\ref{fig:modelsize hotels}.

\section{Instructions with the Multiple Choice Formulation}\label{sec:input_example_mc}

\begin{figure}[!t]
    \centering
    \includegraphics[width=0.49\textwidth]{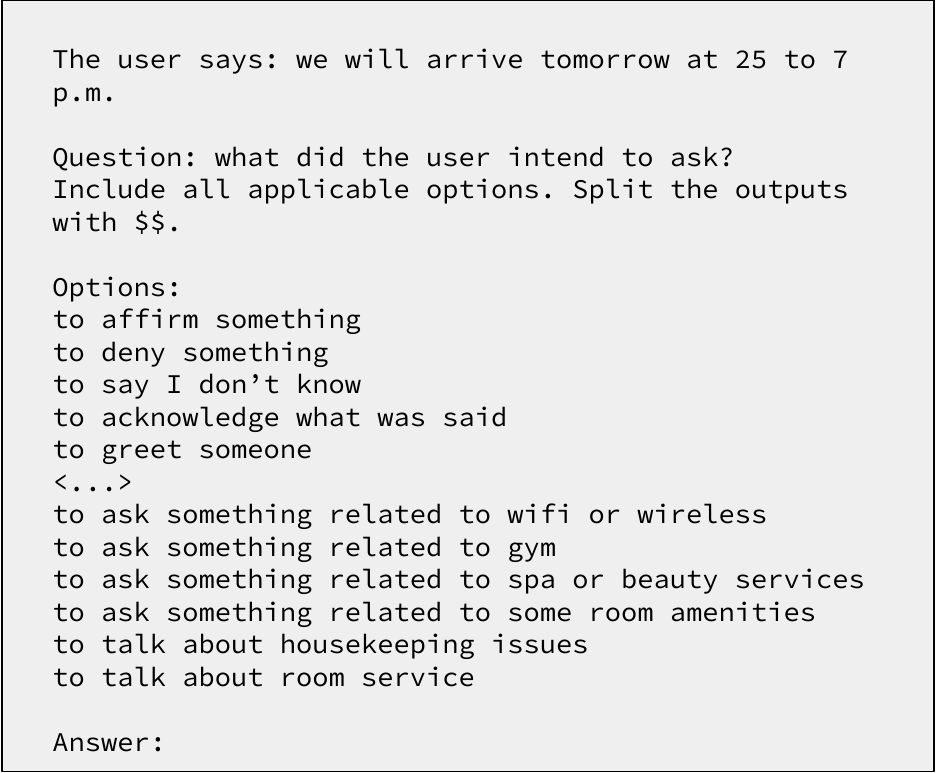}
    \caption{Input example for the multiple-choice formulation in the ID task.}
    \label{fig:input_example_mc}
    \vspace{-3mm}
\end{figure}

Figure~\ref{fig:input_example_mc} shows an example of the multiple choice formulation for the ID task, including the instruction text, user query example and all possible options for the answers.

\end{document}